\documentclass{article}
\usepackage{ijcai18}

\usepackage{times}
\usepackage{xcolor}
\usepackage{soul}
\usepackage[utf8]{inputenc}
\usepackage[small]{caption}
\usepackage{amsmath}
\usepackage{amssymb}
\usepackage{graphicx}
\usepackage{subfigure}
\usepackage{amsfonts}
\usepackage{multirow}
\usepackage{caption}
\usepackage{epsfig}
\usepackage{bm}

\title{Joint CS-MRI Reconstruction and Segmentation with a Unified Deep Network}

\author{Liyan Sun $^\dagger$, Zhiwen Fan $^\dagger$, Yue Huang, Xinghao Ding $^\star$, John Paisley $^\ddagger$\\
$$Fujian   Key   Laboratory   of   Sensing   and   Computing   for   Smart   City,  Xiamen   University, Fujian, China\\
$^\dagger$ The co-first authors contributed equally.\\
$^\star$ Correspondence: dxh@xmu.edu.cn.\\
$^{\ddagger}$Department of Electrical Engineering, Columbia University, New York, NY, USA}

\begin{document}

\maketitle

\begin{abstract}
   The need for fast acquisition and automatic analysis of MRI data is growing in the age of big data. Although compressed sensing magnetic resonance imaging (CS-MRI) has been studied to accelerate MRI by reducing k-space measurements, in current CS-MRI techniques MRI applications such as segmentation are overlooked when doing image reconstruction. In this paper, we test the utility of CS-MRI methods in automatic segmentation models and propose a unified deep neural network architecture called SegNetMRI which we apply to the combined CS-MRI reconstruction and segmentation problem. SegNetMRI is built upon a MRI reconstruction network with multiple cascaded blocks each containing an encoder-decoder unit and a data fidelity unit, and MRI segmentation networks having the same encoder-decoder structure. The two subnetworks are pre-trained and fine-tuned with shared reconstruction encoders. The outputs are merged into the final segmentation. Our experiments show that SegNetMRI can improve {both} the
reconstruction and segmentation performance when using compressive measurements.
% We also discuss broader applications beyond CS-MRI.
\end{abstract}

\section{Introduction}

% \begin{figure}
% \begin{center}
%    {\label {figure1} \includegraphics[width=0.5\textwidth]{Figure1.pdf}}
%    \caption{The integration of MRI reconstruction and segmentation needs to be studied.}
% \label {figure1}
% \end{center}
% \end{figure}

Magnetic resonance imaging (MRI) is an important technique for visualizing human tissue. The raw measurements come in the form of Fourier transform coefficients in ``k-space'' and the MRI can be viewed after an inverse 2D Fourier transform of the fully sampled k-space. Conventionally, radiologists view MRI for diagnosis. However, in areas where medical expertise may be lacking or not sufficient to meet demand, automated methods may also be useful. To this end, automatic MR image segmentation is essential because it allows for finer localization of focus. To take brain segmentation for example, usually four structures emerge including background, gray matter (GM), white matter (WM) and cerebrospinal fluid (CSF). Lesions appearing in white matter are closely associated with various issues such as strokes and Alzheimer's disease \cite{6}.

Although rich anatomical information can be provided by MRI, it is limited by a long imaging period. This can introduce motion artifacts caused by movement of the patient \cite{2} or induce psychological pressures brought by claustrophobia \cite{1}. Thus accelerating imaging speed while maintaining high imaging quality is key for MRI. Compressed sensing (CS) theory \cite{3,4}, which shows the possibility of recovering signals with sub-Nyquist sampling rates, has been introduced to the field of MRI to accelerate imaging. In 2017, the US Food and Drug Administration (FDA) approved CS-MRI techniques for use by two major MRI vendors: Siemens and GE \cite{5}. Thus, one can expect increasing deployment of CS-MRI technique in the future for real-world applications.

To our knowledge, current segmentation algorithms for MRI assume a ``clean'' (i.e., fully-sampled) image as input and do not take CS-MRI scenarios into consideration. Likewise, CS-MRI reconstruction methods do not consider their output's potential downstream segmentation. Although experienced human experts can make relatively robust decisions with CS-reconstructed images, the anticipated increase in the number of CS-reconstructed MRI for clinical application will call for automatic segmentation algorithms optimized for this type of data. Therefore, a unified approach to MRI reconstruction/segmentation under the compressed sensing framework is worthy of exploration.

In this paper, we develop a unified deep neural network called SegNetMRI for joint MRI reconstruction and segmentation with compressive measurements. We build SegNetMRI on two networks: an MRI reconstruction network (MRN) and MRI segmentation network (MSN). The MSN is an encoder-decoder structure network and SegNetMRI is made up of basic blocks which consists of encoder-decoder and data fidelity units. The MRN is pre-trained with pairs of artificially under-sampled and their corresponding fully-sampled MRI and the MSN with fully-sampled MRI and corresponding segmentation labels. We fine-tune the resulting unified network with MSN and MRN sharing the encoder component.
With the basic features produced by the fine-tuned encoder, the MRI reconstruction and segmentation can regularize each other this way.
The outputs are merged by $1\times1$ convolution.
% Throughout the paper, we only focus on brain MRI data reconstruction and segmentation.

% Our contributions can be summarized as follows:
% \begin{itemize}
%   \item To our knowledge, for the first time, we prove the segmentation performance of the well-trained MRI model on the reconstructed MR images produced by popular CS-MRI models still needs large improvement by experiments.
%   \item We propose an SegNetMRI architecture to jointly reconstruct and segment CS-MRI data. The information in mid-level segmentation can guide the reconstruction and the reconstruction can adapt to the segmentation in turn.
%   \item We show the MR image reconstruction and segmentation are positively associated with no need to sacrifice any of them. The proposed SegNetMRI model can improve both the reconstruction quality and segmentation accuracy. Besides, we open the discussion on building a unified system to incorporate visual tasks in different levels like medical image reconstruction and segmentation.
% \end{itemize}

\section{Background and Related Work}

\paragraph{MRI segmentation}
Broadly speaking, the research in MRI segmentation can be categorized into three classes: (1) atlas-based segmentation with registration; (2) machine learning models with hand-crafted features; (3) deep learning models. Atlas-based segmentation \cite{20,21} requires accurate registration and is time-consuming, so it is impractical in real applications that require fast speed. In the second class, manually designed features are fed into classifiers, e.g., 3D Haar/spatial features into random forests \cite{35}. These hand-crafted features are not very flexible in encode diverse patterns within MRI data. Recently deep learning based models have been propose,d such as a 2D convolutional neural network \cite{23,24}, a 3D convolutional neural network \cite{25,26}, and parallelized long short-term memory (LSTM) \cite{27}. They can learn semantic image features from data, leading to the state-of-the-art performance in MRI segmentation. However, these MRI segmentation models
do not take the input quality into consideration, but assume full measurements.

\paragraph{Compressed Sensing MRI}
We denote the underlying vectorized MRI $x \in {\mathbb C^{P \times 1}}$ which we seek to reconstruct from the sub-sampled vectorized k-space data $y\in {\mathbb C^{Q \times 1}}$ ($Q \ll P$). CS-MRI is then typically formulated as
\begin{equation}\label{eq1}
      x = \arg \min \left\| {{F_u}x - y} \right\|_2^2 + {f_\theta }\left( x \right),
\end{equation}
where the $F_u\in {\mathbb C^{Q \times P}}$ denotes the under-sampled Fourier matrix. The $\ell_2$ term is the data fidelity and ${f_\theta }\left( \cdot \right)$ represents a regularization with parameter $\theta$ to constrain the solution space.

The main research focus of CS-MRI is proposing better $f_\theta$ and efficient optimization techniques. In the first CS-MRI work called SparseMRI \cite{8}, wavelet domain $\ell_1$ sparsity plus image total variation are imposed as regularizations. More complicated wavelet variants are designed for CS-MRI in PANO \cite{12} and GBRWT \cite{13}. Dictionary learning techniques are also introduced in CS-MRI, e.g., DLMRI \cite{14} and BPTV \cite{16}. These works can all be categorized as sparsity-based CS-MRI methods; they model the MRI with a ``shallow'' sparse prior, which often tends to over-smooth the image.

Recently, deep neural networks have been introduced to CS-MRI. Researchers have directly applied convolutional neural networks (CNN) to learn a direct mapping from the zero-filled MRI $F_u^Hy$ (obtained by zero-padding the unsampled positions in k-space) to the true MRI \cite{17}. A deep residual architecture was also proposed for this same mapping \cite{18}. Data fidelity terms have been incorporated into the deep neural network by \cite{19} to add more guidance. These deep learning based CS-MRI models have achieved higher reconstruction quality and faster reconstruction speed.
% Furthermore, the high-level semantic modeling ability of deep neural network has greater potential of merging more information from mid or high level task into the reconstruction network over the sparsity based CS-MRI methods, although few literatures are devoted to discussing on it.

\paragraph{Combining visual tasks}
The combination of different visual tasks in a unified model has been investigated in the field of computer vision. For example, a joint blind image restoration and recognition model \cite{28} based sparse coding was proposed for face recognition with low-quality images. In the image dehazing model AOD-Net \cite{32}, the researchers discuss detection in the presence of haze by performing dehazing during detection. In the MRI field, 3T-obtained images have been used in joint segmentation and super-resolution (7T) image generation \cite{37}.

% In a recent work in preprint \cite{33}, the low-level visual task image denoising is combined with image segmentation/classification is dealt in a single deep architecture, where the denoising network and segmentation network are cascaded. Only the denoising network is updated and the already well-trained segmentation network keeps fixed during the fine-tuning. The deep architecture can significantly improve the segmentation accuracy. It is worthy noting the denoised images by the model contain more visual details than the single denoising network, but peak signal-to-noise ratio (PSNR) index value of the joint architecture is lower, which indicating greater absolute reconstruction errors or even ``fake" details are produced. This is unfavorable for medical image reconstruction because information lossless is a primary goal for providing liable diagnosis.

\begin{figure}
\begin{center}
   {\label {figure2} \includegraphics[width=1\columnwidth]{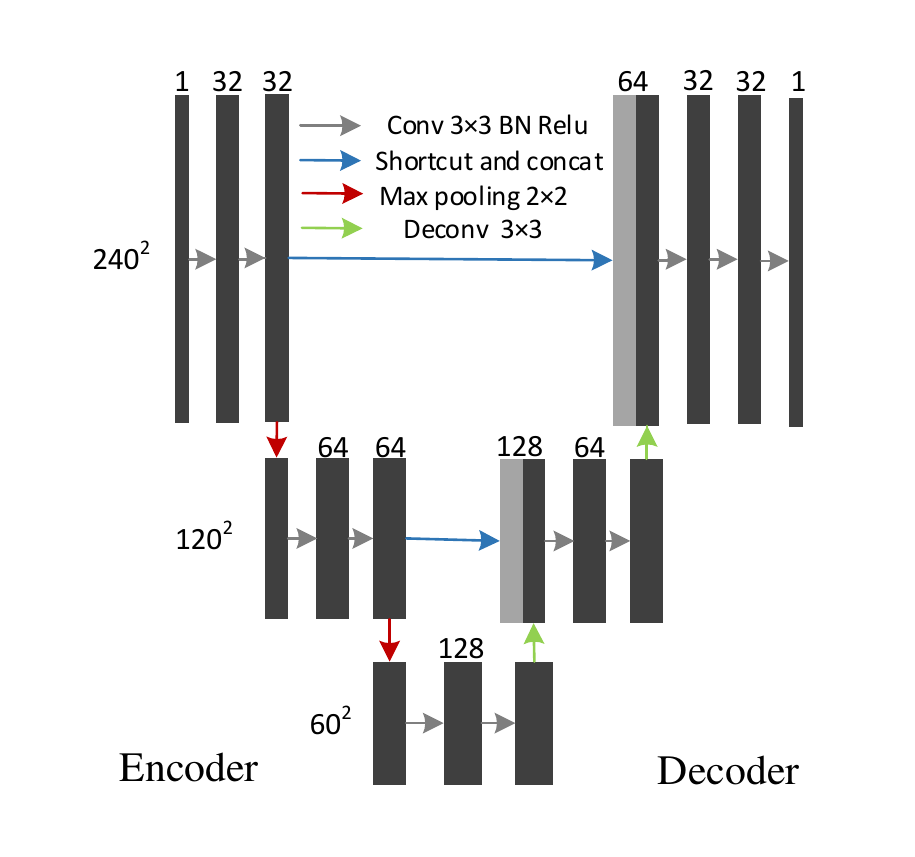}}
   \caption{The MSN architecture composed of an encoder and decoder. It is used to assess the segmentation accuracy on different reconstructed MRI produced by various CS-MRI methods.}
\label {figure2}
\end{center}
\end{figure}

\section{Methodology}

In this section, we give a detailed description of the proposed SegNetMRI model. First, we propose a segmentation network baseline and test popular CS-MRI methods on the well-trained model. Next we propose a MRI reconstruction network formed by cascading basic blocks. We show the proposed MRI reconstruction network achieves better performance on segmentation over conventional sparsity based CS-MRI models, but still inferior to the full-sampled MR image. We further propose the SegNetMRI model to merge the MRI reconstruction and segmentation into an single model.

\begin{figure*}
\centering
\subfigure[ ZF] {\label {figure3a} \includegraphics[width=0.187\textwidth]{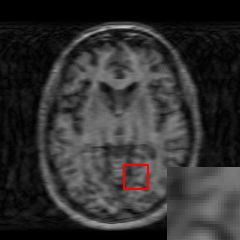}}
   \subfigure[ PANO] {\label {figure3b} \includegraphics[width=0.187\textwidth]{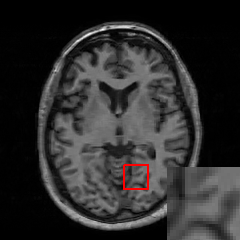}}
   \subfigure[ MRN$_{5}$] {\label {figure3c} \includegraphics[width=0.187\textwidth]{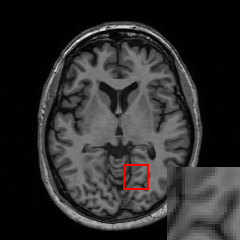}}
   \subfigure[ Full-sample (GT) ] {\label {figure3d} \includegraphics[width=0.187\textwidth]{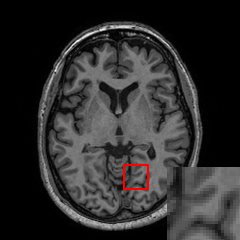}}
   \subfigure[ Mask] {\label {figure3e} \includegraphics[width=0.187\textwidth]{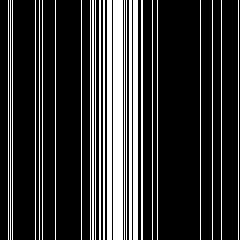}}\\
   \subfigure[ ZF Seg] {\label {figure3f} \includegraphics[width=0.187\textwidth]{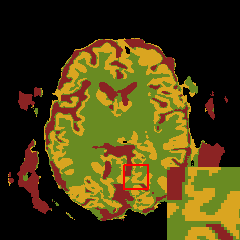}}
   \subfigure[ PANO Seg] {\label {figure3g} \includegraphics[width=0.187\textwidth]{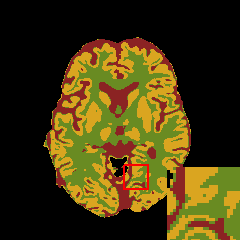}}
   \subfigure[ MRN$_{5}$ Seg] {\label {figure3h} \includegraphics[width=0.187\textwidth]{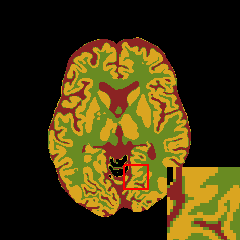}}
   \subfigure[ Full-sample Seg] {\label {figure3i} \includegraphics[width=0.187\textwidth]{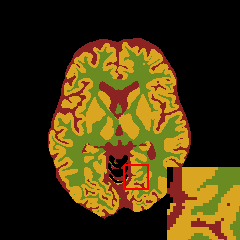}}
   \subfigure[ Manual Seg] {\label {figure3j} \includegraphics[width=0.187\textwidth]{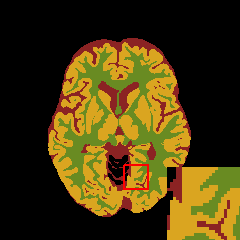}}
   \caption{First row: The reconstructed MRI using different CS-MRI methods and a $20\%$ sampling mask. These are then segmented by an independently-trained segmentation model based on the state-of-the-art U-Net (referred to as MSN in this paper for its MRI application).  }
\label {figure3}
\end{figure*}

\subsection{Illustration: Segmentation after CS-MRI}

We first test several popular CS-MRI outputs on an automatic MRI segmentation model to assess the impact of compressed sensing for this task. Inspired by the state-of-the-art medical image segmentation model U-Net \cite{34}, we propose the MRI segmentation network (MSN) shown in Figure \ref{figure2}.  The segmentation encoder (SE) component, using convolution and pooling, can extract features from the input image at different scales, and the segmentation decoder (SD) component, using a deconvolution operation, predicts the the pixel-wise segmentation class from these features. Shortcut connections are used by this model to directly send lower-layer features to high-layer features by concatenation.

Training this MSN model using fully-sampled MRI and their segmentation labels, we test this models performance on reconstructed MRI produced by various CS-MRI methods. We use a $20\%$ Cartesian under-sampling mask as shown in Figure \ref{figure3e}. Our tested methods including the degraded zero-filled (ZF) reconstruction as baseline, PANO \cite{12} and the proposed MRN model which will be discussed in the following section\footnote{We adjust the parameters of PANO for this problem.}.

We observe that the zero-filled (ZF) reconstruction in Figure \ref{figure3a} produces a low-quality MR image, which leads to the poor segmentation performance shown in Figure \ref{figure3f}. The PANO reconstructed MRI in Figure \ref{figure3b} has been segmented better in Figure \ref{figure3g}, but is still far from satisfactory because of the loss of structural details in the reconstruction. The segmentation using the fully-sampled (FS) MRI in Figure \ref{figure3d} is shown in Figure \ref{figure3i}. Though this isn't the ground truth segmentation, it is the segmentation performed on the ground truth MRI, and so represented an upper bound for CS-MRI on this segmentation task. The manually label segmentation is shown in Figure \ref{figure3j}. This experiment shows that while CS-MRI can substantially improve the reconstruction quality visually, the fine structural details which are important for segmentation can still be mission, leaving much space for further improvement.

\begin{figure*}
\begin{center}
%   \subfigure[The MRN$_N$ architecture.] {\label {figure4a} \includegraphics[width=1\textwidth]{MReconNet.pdf}}
%   \subfigure[The proposed UniNet$_N$ architecture.] {\label {figure4b} \includegraphics[width=1.027\textwidth]{UniNet.pdf}}
   {\includegraphics[width=1\textwidth]{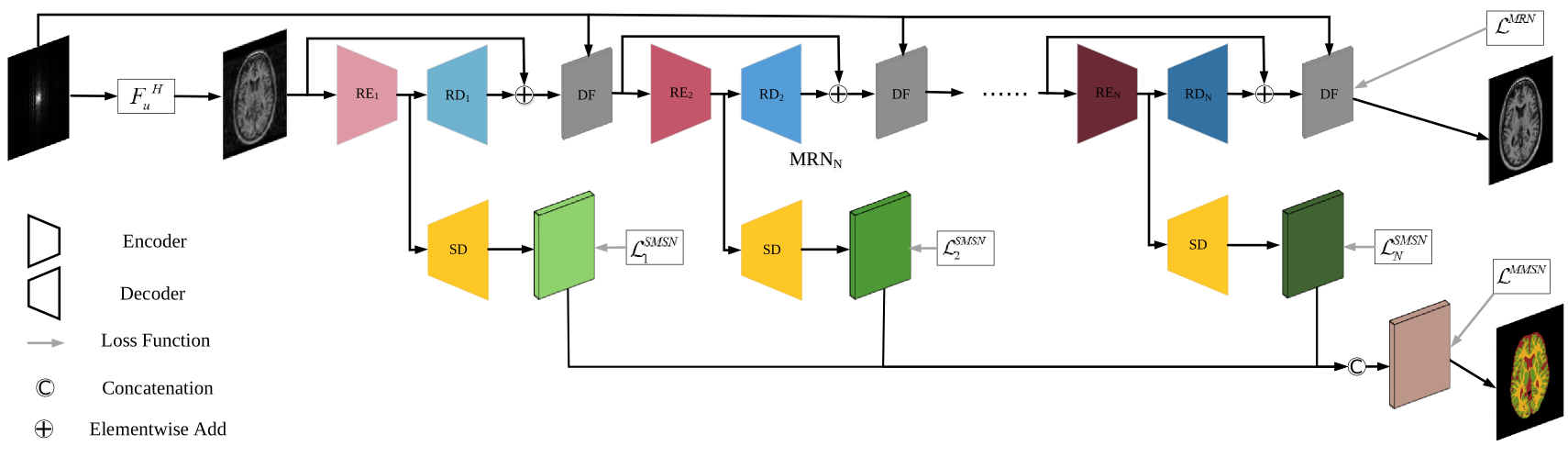}}
   \caption{The SegNetMRI structure, formed by connecting the discussed MRN (top) for reconstruction, with MSN (bottom) for segmentation.}
\label {figure4}
\end{center}
\end{figure*}

\subsection{An MRI Reconstruction Network}

Deep learning for CS-MRI has the advantage of large modeling capacity, fast running speed, and high-level semantic modeling ability, which eases the integration of high-level task information compared with traditional sparsity-based CS-MRI models. Therefore, we adopt the same MSN encoder-decoder architecture from Figure \ref{figure2} as a basic encoder-decoder unit with a global residual shortcut, which has been proven to help network training. The encoder-decoder unit produces the reconstructed MRI. Since with deep neural networks the information loss can become severe, we introduce a data fidelity (DF) unit to help correct the Fourier coefficients of the reconstructed MRI produced by the encoder-decoder architecture on the sampled positions in k-space. This takes advantage of the fact that we have accurate measurements at the under-sampled k-space locations, and so the layers of the network should not override this information. (The details of the data fidelity unit can be found in \cite{19}.)

The encoder-decoder architecture and data fidelity unit make up a basic block. As more blocks are stacked in a cascaded manner, the quality of the reconstructed MRI of each block can be gradually improved \cite{19}. We therefore cascade $N$ such basic units to form the MRI reconstruction network (MRN$_N$) in Figure \ref{figure4}. The reconstruction encoder in different blocks extract features at different depth. Previously in Figure \ref{figure2}, we observed that MRN$_5$ achieves better reconstruction performance in Figure \ref{figure3c} than the non-deep PANO method, but the segmentation output in Figure \ref{figure3h} (MRN$\rightarrow$MSN) is still inferior to the fully-sampled segmentation. This motivates the following joint framework.

\subsection{SegNetMRI: A Unified Deep Neural Network}

Based on these observations, we propose a joint framework for CS-MRI reconstruction/segmentation that uses a deep neural network. We call this joint network SegNetMRI, which is shown in Figure \ref{figure4}.

To learn this model, we first pre-train separate models. We pre-train the MRN$_N$ with under-sampled and fully-sampled MRI training pairs. Similarly, we pre-train the MSN with fully-sampled MRI and their corresponding segmentation labels. After training MSN, we leave out the encoder component (SE) and keep the decoder component (SD). We then connect the single decoder component of the MSN (SD) to each of the encoder components of each MRN (called RE$_n$) within each block. The resulting $N$ outputs of the MSN decoder for each block are concatenated and merged to give the final segmentation via a $1\times1$ convolution. After pre-training separately and initializing the remaining parts, the parameters of SegNetMRI$_N$ (with $N$ blocks in the MRN portion, but a single segmentation decoder duplicated $N$ times) are then fine-tuned. Therefore, both the reconstruction and segmentation tasks share the \textit{same} encoders, but have \textit{separate} decoders for their respective tasks.

The rationale for this architecture is the following:
\begin{enumerate}
 \item With the pre-training of MRN$_N$, the reconstruction encoder extracts basic features in different blocks. In SegNetMRI, the sharing of the reconstruction encoders between MRN and MSN means that these same features are used for both reconstruction and segmentation, which can help the two problems regularize each other.
 \item The segmentation component uses information at various depths in the cascaded MRN, and combines this information in the decoder. The $1\times1$ convolution used to merge the outputs of the segmentation decoder at each layer can be viewed as ensemble learning.
\end{enumerate}

% In related work of \cite{33}, a proposed denoising network and segmentation network are also organized in a cascaded way for the denoising/segmentation problem. Although the denoising network is updated by the gradient flowing through the fixed segmentation network, the gradient vanish effect may mislead the denoising network in undesirable optimization direction, the resulting information loss may degrade the reconstruction quality or even in turn limit the improvement of segmentation performance. On the contrary, in the proposed SegNetMRI architecture, we first extract the basic features of the under-sampled MRI data by the reconstruction encoder in each block of the MRN model. With the fine-tuned basic features produced by the sharing encoder fed to the decoder of MRN and MSN, the MRN and MSN regularize each other in much more fundamental way with less information loss.

\paragraph{Loss function}
We adopt the $\ell_2$ Euclidean distance as the loss function for pre-training the MRN,
\begin{equation}\label{eq2}
  {\mathcal{L}^{{\rm{MRN}}}} = \frac{1}{L}\sum\nolimits_{i = 1}^L {\left\| {x_i^{fs} - {x_i}} \right\|_2^2},
\end{equation}
where $x_i^{fs}$ and $x_i$ are the $i^{th}$ full-sampled training image and the output of the MRN, respectively. To pre-train the MSN, we adopt the pixel-wise cross-entropy loss function
\begin{equation}\label{eq3}
  {\mathcal{L}^{{\rm{MSN}}}} =  - \sum\nolimits_{i = 1}^L {\sum\nolimits_{j = 1}^N {\sum\nolimits_{c = 1}^C {t_{ijc}^{gt}} } } \ln {t_{ijc}},
\end{equation}
where $C$ tissue classes are to be classified. $t^{gt}$ and $t$ is the pixel-wise ground-truth label and the MSN predicted label, respectively.

Pre-training the MRN and MSN this way, we then construct and fine-tune SegNetMRI with the following loss function
\begin{equation}\label{eq4}
{{\cal L}^{{\rm{SegNetMRI}}}} = {{\cal L}^{{\rm{MRN}}}} + \lambda {{\cal L}^{{\rm{OMSN}}}}.
\end{equation}
We set  $\lambda = 0.01$ in our experiments. Note the overall MSN (OMSN) loss containing $N+1$ loss function terms if the SegNetMRI contains $N$ blocks is
\begin{equation}\label{eq5}
{\mathcal{L}^{{\rm{OMSN}}}} = \frac{1}{{N + 1}}\left( {{\mathcal{L}^{{\rm{MMSN}}}} + \sum\nolimits_{i = 1}^N {{\mathcal{L}_i}^{{\rm{SMSN}}}} } \right),
\end{equation}
where the ${\cal L^{{\rm{MMSN}}}}$ is the loss for the merged prediction and ${\cal L^{{\rm{SMSN}}}}$ is the loss for each sub-MSN decoder prediction.

\section{Experiments and Discussions}

\begin{figure*}
\begin{center}
   \subfigure[ GBRWT+MSN]        {\label {figure5a} \includegraphics[width=0.155\textwidth]{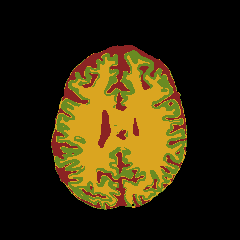}}
   \subfigure[ MRN$_5$+MSN]      {\label {figure5b} \includegraphics[width=0.155\textwidth]{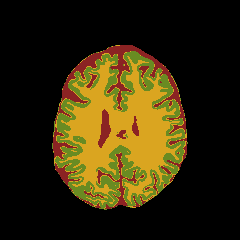}}
   \subfigure[ \protect\cite{33}]        {\label {figure5c} \includegraphics[width=0.155\textwidth]{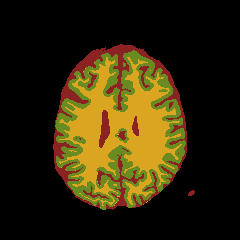}}
   \subfigure[ SegNetMRI$_5$]       {\label {figure5d} \includegraphics[width=0.155\textwidth]{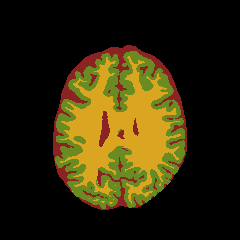}}
   \subfigure[ FS+MSN]           {\label {figure5e} \includegraphics[width=0.155\textwidth]{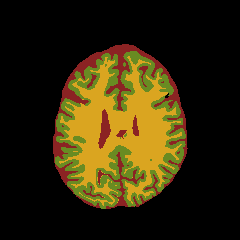}}
   \subfigure[ GroundTruth]      {\label {figure5f} \includegraphics[width=0.155\textwidth]{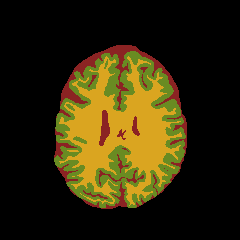}}
   \caption{The segmentations of the compared methods. }
\label {figure5}
\end{center}
\end{figure*}

\begin{table*}[]
\centering
\begin{tabular}{|l|c|l|l|c|l|l|c|c|l|}
\hline
\multirow{2}{*}{Methods} & \multicolumn{3}{c|}{GM}                                      & \multicolumn{3}{c|}{WM}                                      & \multicolumn{3}{c|}{CSF}                                      \\ \cline{2-10}
                         & DC    & \multicolumn{1}{c|}{HD}   & \multicolumn{1}{c|}{AVD} & DC    & \multicolumn{1}{c|}{HD}   & \multicolumn{1}{c|}{AVD} & DC    & HD                        & \multicolumn{1}{c|}{AVD}  \\ \hline
GBRWT+MSN                & 75.55 & 2.24                      & 4.21                     & 65.56 & 1.90                      & 3.10                     & 76.50 & 1.77                      & 2.69                      \\ \hline
MRN$_5$+MSN                  & 79.36 & 2.06                      & 3.57                     & 65.76 & \multicolumn{1}{c|}{1.88} & 2.96                     & 78.43 & \multicolumn{1}{l|}{1.64} & \multicolumn{1}{c|}{2.33} \\ \hline
\cite{33}                    & 83.41 & 1.81                      & 2.96                     & 78.05 & 1.24                      & 1.61                     & 77.81 & 1.76                      & 2.58                      \\ \hline
SegNetMRI$_5$                   & \textbf{86.38} & \multicolumn{1}{c|}{\textbf{1.66}} & \textbf{2.52}                     & \textbf{81.49} & \textbf{1.08}                      & \textbf{1.34}                     & \textbf{79.23} & \multicolumn{1}{l|}{\textbf{1.61}} & \textbf{2.23}                      \\ \hline
FS+MSN                   & 87.36 & 1.60                      & 2.33                     & 85.94 & 1.00                      & 1.14                     & 81.01 & 1.61                      & 2.18                      \\ \hline
\end{tabular}
\caption{The segmentation comparison of the different models using DC (\%), HD and AVD (\%) as metrics. FS+MSN is the segmentation performance when the ground truth MRI is known.}
\label{SegTable}
\end{table*}

\subsection{Implementation Details}

\paragraph{Setup}
We implement all deep models on TensorFlow for the Python environment using a NVIDIA Geforce GTX 1080Ti with 11GB GPU memory and Intel Xeon CPU E5-2683 at 2.00GHz. We show the hyperparameter settings of encoder-decoder architecture used for both MRN and MSN in Figure \ref{figure2}. We use batch normalization to stabilize training. ReLU is used as activation function except for the last convolution layer of the encoder-decoder unit within each MRN block, where the identity map is applied for residual learning. We apply Xavier initialization for pre-training MRN and MSN. MSN is pre-trained for $60,000$ iterations using $64\times64$ fully-sampled MRI patches randomly cropped ($16$ patches in a batch) and MRN is pre-trained for $30,000$ iterations using the entire training image ($4$ images in a batch). We then fine-tune the SegNetMRI model for $8,000$ further iterations using entire images ($4$ images in a batch). ADAM is chosen as optimizer. We select the initial learning rate to be $0.0005$, the first-order momentum to be $0.9$ and the second momentum to be $0.999$.

\paragraph{Data}
We test our SegNetMRI architecture on the MRBrainS datasets from the Grand Challenge on MR Brain Image Segmentation (MRBrainS) Workshop \cite{7}. The datasets are acquired using 3T MRI scans. Five datasets are provided containing T1-1mm, T1, T1-IR and T2-FLAIR imaging modalities already registered and with manual segmentations. Here we use the T1 MRI data of the size $240\times240$ throughout the paper. We use four datasets for training (total 192 slices) and one dataset for testing (total 48 slices). We adopt the same data augmentation technique discussed in \cite{36}.

\begin{figure}[ht!]
\begin{center}
   \subfigure[ ZF]                 {\label {figure6a} \includegraphics[width=0.15\textwidth]{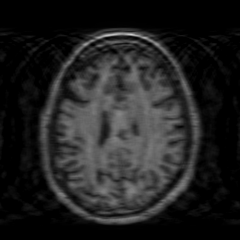}}
   \subfigure[ GBRWT]              {\label {figure6b} \includegraphics[width=0.15\textwidth]{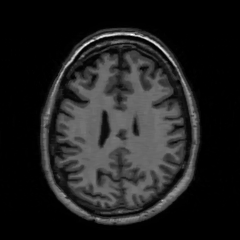}}
   \subfigure[ MRN$_5$]            {\label {figure6c} \includegraphics[width=0.15\textwidth]{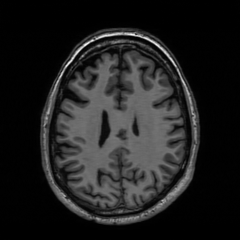}}
   \subfigure[ Huang]              {\label {figure6d} \includegraphics[width=0.15\textwidth]{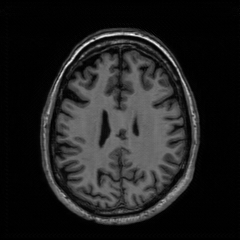}}
   \subfigure[ SegNetMRI$_5$]         {\label {figure6e} \includegraphics[width=0.15\textwidth]{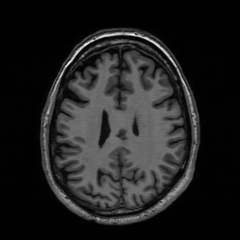}}
   \subfigure[ FS]                 {\label {figure6f} \includegraphics[width=0.15\textwidth]{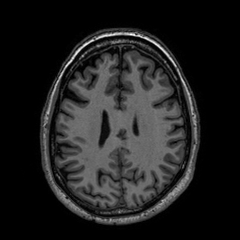}}
   \subfigure[\tiny GBRWT Error]        {\label {figure6g} \includegraphics[width=0.11\textwidth]{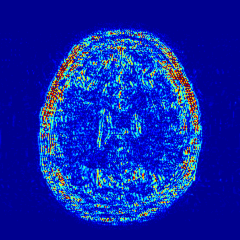}}
   \subfigure[\tiny MRN$_5$ Error]      {\label {figure6h} \includegraphics[width=0.11\textwidth]{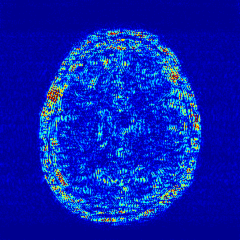}}
   \subfigure[\tiny Huang Error]        {\label {figure6i} \includegraphics[width=0.11\textwidth]{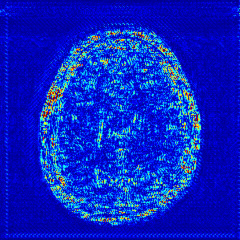}}
   \subfigure[\tiny SegNetMRI$_5$ Error]   {\label {figure6j} \includegraphics[width=0.11\textwidth]{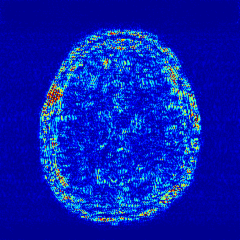}}
   \caption{The reconstructed MR images with different CS-MRI methods in the first two rows and the residual maps in the third row. }
\label {figure6}
\end{center}
\end{figure}

\subsection{Experimental Results}

To evaluate the segmentation performance, we compare the proposed SegNetMRI$_5$ ($5$ blocks) with the inputting a fully-sampled MRI in MSN (FS+MSN) as ground truth performance, inputing the MRN$_5$ reconstruction in MSN (MRN$_5$+MSN) -- i.e., no fine-tuning), inputting GBRWT reconstruction in MSN (GBRWT+MSN). GBRWT \cite{13} represents the state-of-the-art performance in the conventional sparse based CS-MRI methods. Finally, we also compare with the joint framework proposed by \cite{33}, where only the reconstruction network is fine-tuned in MRN$_5$+MSN using the loss function ${\mathcal{L}^{{\rm{MRN}}}} + \lambda{\mathcal{L}^{{\rm{MSN}}}}$.
%using the same loss function in Equation \ref{eq4}.
The same under-sampling mask shown in Figure \ref{figure3e} is again used.

We show qualitative comparison in Figure \ref{figure5}. We colorize segmentation corresponding white matter, gray matter and cerebrospinal fluid with yellow, green and red. We observe that the proposed SegNetMRI$_5$ model provides better segmentation and approximates the ideal FS+MSN most closely, both of which are not perfect compared with the human labeling. For quantitative evaluation, we use the Dice Coefficient (DC), the $95$th-percentile of the Hausdoff distance (HD) and the absolute volume difference (AVD), as also used in the MRBrainS challenge \cite{7}. Larger DC, and smaller HD and AVD, indicate better segmentation. We show these results in Table \ref{SegTable}, which is consistent with the subjective evaluation.

In addition to the improvement of segmentation accuracy, we also evaluate the reconstruction quality of the SegNetMRI model. We show the reconstructed MRI from SegNetMRI$_5$, the model in \cite{33}, MRN$_5$ and GBRWT in Figure \ref{figure6}, along with their corresponding residuals. (We set the error ranges from $[0~~0.1]$ on a $[0~~1]$ pre-scaled image.) We find that SegNetMRI achieves the minimal reconstruction error, especially in the meaningful tissue regions. We also give averaged reconstruction performance measures in Table \ref{ReconTable} using peak signal-to-noise ratio (PSNR) and the corresponding normalized mean squared error (NMSE) on all 37 test MRI.

\begin{table}[t]
\centering
\small
\begin{tabular}{|l|c|c|c|c|}
\hline
& GBRWT  & MRN$_5$    & \cite{33}  & SegNetMRI$_5$ \\ \hline
PSNR    & 31.80  & 33.94  & 33.47  & \textbf{34.27}  \\ \hline
NMSE    & 0.0584 & 0.0361 & 0.0388 & \textbf{0.0333} \\ \hline
\end{tabular}
\caption{The averaged PSNR (dB) and NMSE on 37 test MRI data.}
\label{ReconTable}
\end{table}

\paragraph{Discussion}
It is worth noting that the the model in \cite{33} achieves better segmentation performance than the MRN$_5$ model, but the reconstruction quality is worse than MRN$_5$ both qualitatively and quantitatively. The original work in \cite{33} is devote to joint natural image denoising and segmentation, while for medical image analysis, the absolute reconstruction and segmentation accuracy is equally important. Usually, segmented MRI will also be provided by radiologists for diagnosis. A reconstruction error can thus cause the loss of valuable diagnostic information. In contrast, the SegNetMRI model achieves better performance on both reconstruction and segmentation.

In SegNetMRI$_N$, the output of $N$ MSN decoders are concatenated and merged into the final segmentation using a $1\times1$ convolution. The ensemble learning can make full use of the information from different depth of the SegNetMRI and produce better segmentation accuracy. We take the SegNetMRI$_5$ on the gray matter tissue of all the test MRI data for example. In Table \ref{Merge}, we show the segmentation performance of the outputs from each block in SegNetMRI$_5$ model without the $1\times1$ convolution, and we compare them with the segmentation output produced after merging the SegNetMRI$_5$ outputs with $1\times1$ convolution. The output of SegNetMRI$_5$ achieves better segmentation performance, showing the effectiveness of this ensemble structure.

In Figure \ref{figure7}, we show the segmentation accuracy (in Dice Coefficient metric) as a function of blocks, $N$. The reconstruction quality (in PSNR metric) improves as the number of the blocks increases in the SegNetMRI model, but at the expense of longer training time.

\begin{table}[t]
\centering
\begin{tabular}{|l|c|c|c|c|c|c|}
\hline
GM  & B$_1$    & B$_2$    & B$_3$    & B$_4$    & B$_5$    & Merged \\ \hline
DC  & 75.15 & 80.31 & 83.64 & 81.02 & 85.66 & \textbf{86.38}  \\ \hline
HD  & 2.15  & 1.95  & 1.77  & 1.90  & 1.68  & \textbf{1.66}   \\ \hline
AVD & 4.35  & 3.58  & 2.90  & 3.43  & 2.60  & \textbf{2.52}   \\ \hline
\end{tabular}
\caption{We compare the segmentation performance of the outputs from each block in SegNetMRI$5$ model without the $1\times1$ convolution and the segmentation output produced by the SegNetMRI$_5$ via merging.}
\label{Merge}
\end{table}

\begin{figure}
\begin{center}
   {\label {figure7} \includegraphics[width=0.45\textwidth]{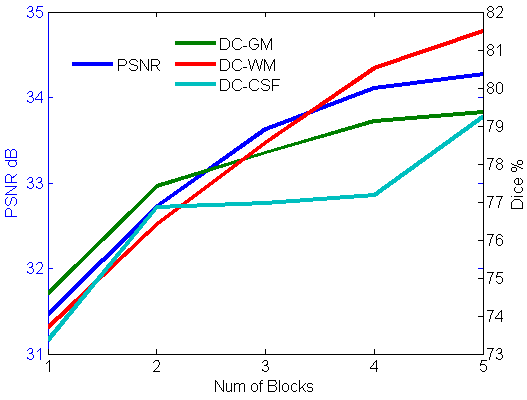}}
   \caption{The model performance of SegNetMRI$_N$ architecture as a function of the number of blocks.}
\label {figure7}
\end{center}
\end{figure}

\section{Conclusion}

Automatic segmentation of MRI is an important problem in medical imaging, and with the recent adoption of CS-MRI by industry, segmentation techniques that take CS-MRI reconstruction into account are needed. After verifying that the two tasks suffer when done independently, we proposed a deep neural network architecture called SegNetMRI to merge the MRI reconstruction and segmentation problems into a joint framework. Our experiments show that doing simultaneous reconstruction and segmentation can positively reinforce each other, improving both tasks significantly.

\section*{Acknowledgement}
\thanks{This work was supported in part by the National Natural Science Foundation of China under Grants 61571382, 81671766, 61571005, 81671674, U1605252, 61671309 in part by the Guangdong Natural Science Foundation under Grant 2015A030313007, in part by the Fundamental Research Funds for the Central Universities under Grant 20720160075, 20720180059, in part by the National Natural Science Foundation of Fujian Province, China under Grant 2017J01126. (Corresponding author: Xinghao Ding)}

\bibliographystyle{named}
\bibliography{ijcai18}

\end{document}